\documentclass[conference]{IEEEtran}
\IEEEoverridecommandlockouts
\usepackage{cite}
\usepackage{amsmath,amssymb,amsfonts}
\usepackage{algorithmic}
\usepackage{graphicx}
\usepackage{textcomp}
\usepackage{xcolor}
\def\BibTeX{{\rm B\kern-.05em{\sc i\kern-.025em b}\kern-.08em
    T\kern-.1667em\lower.7ex\hbox{E}\kern-.125emX}}

\usepackage{fancyhdr}
\usepackage{subcaption}
\usepackage{enumitem}
\usepackage{hyperref}
\begin{document}

\title{Expresso-AI: Explainable Video-Based Deep Learning Models for Depression Diagnosis
}

\author{\IEEEauthorblockN{1\textsuperscript{st} Felipe Moreno}
\IEEEauthorblockA{\textit{Media Lab} \\
\textit{MIT}\\
Cambridge, USA \\
pipemon@mit.edu}
\and
\IEEEauthorblockN{2\textsuperscript{nd} Sharifa Alghowinem}
\IEEEauthorblockA{\textit{Media Lab} \\
\textit{MIT}\\
Cambridge, USA \\
sharifah@mit.edu}
\and
\IEEEauthorblockN{3\textsuperscript{rd} Hae Won Park}
\IEEEauthorblockA{\textit{Media Lab} \\
\textit{MIT}\\
Cambridge, USA \\
haewon@media.mit.edu}
\and
\IEEEauthorblockN{4\textsuperscript{th} Cynthia Breazeal}
\IEEEauthorblockA{\textit{Media Lab} \\
\textit{MIT}\\
Cambridge, USA \\
cynthiab@media.mit.edu}
}

\maketitle
\thispagestyle{fancy}

\begin{abstract}
 %intro %importance %motivation 
 % Detecting depression is of paramount importance for society, as early diagnosis and intervention can significantly improve the quality of life for millions of individuals worldwide, ultimately reducing the emotional, social, and economic burden caused by this prevalent mental health condition.
 % 
 Given the widespread prevalence of depression and its consequential impact on individuals and society, it is crucial to obtain objective measures for early diagnosis and intervention. As a multidisciplinary topic, these objective measures should be interpretable and accessible to health care professionals, ensuring effective collaboration and treatment planning in the realm of mental health care.
 %gap
 Even though current automated depression diagnosis approaches have improved over the last decade, a critical gap exists as they often lack affect-specificity and interpretability, limiting their practical application and potential impact on mental health care.
 In particular, interpretability from temporal activities from videos when deep models are used is not fully explored. 
 %RQ
 In this study, we present a novel framework for analyzing Deep Neural Networks' decisions when trained on facial videos, specifically focusing on automatic depression severity diagnosis. 
 %method
 By fine-tuning Deep Convolutional Neural Networks (DCNN) pre-trained on Action Recognition datasets on depression severity facial videos from the AVEC depression dataset, our framework is able to interpret the model's saliency maps by examining face regions and temporal expression semantics. 
 Our approach generates both visual and quantitative explanations for the model's decisions, providing greater insight into its reasoning. 
 %results
 In addition to this interpretability, our video-based modeling has improved upon previous single-face benchmarks for visual depression diagnosis, resulting in enhanced predictive performance. 
 %conclusion
Overall, our work demonstrates the successful development of a framework capable of generating hypotheses from a facial model's decisions while simultaneously improving depression's predictive capabilities.
%future

\end{abstract}


\section{Introduction}
% a paragraph on  %intro, importance, motivation
Depression is a pervasive mental health issue affecting millions of people globally. According to the World Health Organization (WHO), over 264 million people of all ages suffer from depression, making it one of the leading causes of disability worldwide \cite{WHO-depression}. Current diagnostic methods for depression primarily rely on subjective assessments, such as self-reports and clinical interviews, which may be influenced by factors like personal bias, cultural differences, and subjective interpretations. Given the widespread prevalence of depression and its consequential impact on individuals and society, there is a pressing need for developing objective measures to accurately identify and assess depressive symptoms. Such objective measures should be interpretable and accessible to health care professionals, ensuring effective collaboration and treatment planning in the realm of mental health care. By addressing this need, we can pave the way for improved early diagnosis, intervention, and ultimately, better mental health outcomes for those affected by depression.

% a paragraph on  gap
In an effort to address the subjective nature of depression diagnosis, attempts have been made to employ machine learning methods to develop objective measures for diagnosing depression. These automated methods have analyzed various aspects, including video (e.g., facial expressions, head orientations, eye movements), audio, and brain signals, among others \cite{pampouchidou2017automatic}. 
However, even when investigating the same modality, different studies implement different approaches in terms of extracted features, machine learning algorithms, and reported results, making the comparison and conclusion difficult.
Among the studies focusing on automated depression diagnosis (ADD), very few have provided insights or interpretations that could explain the model's performance to human observers. This lack of interpretability is a common issue in machine learning studies, particularly with the rise of deep learning modeling \cite{miller2019explanation}. Interpretability of depression diagnosis modeling is particularly crucial for communicating its performance to clinicians who are interested not only in understanding the dynamics of depression symptoms, but also to be part of the diagnosis decision made by the model.

% a paragraph on  research question (RQ) and summary of the method
The current state of video-based modeling for depression diagnosis lags behind image-based advances in both performance and explainability, where the majority of vision models diagnose depression based on images \cite{pampouchidou2017automatic}. Video modeling, with its convolution capabilities like 3D CNNs, has the potential to extract features from the dynamics of motion, going beyond appearance-based analysis. Despite these benefits, existing explainability literature barely addresses the time dimension in their algorithms, with initial approaches mainly focusing on visualizations and heat-maps for quantitative analysis. Therefore, we developed a framework\footnote{Code is available at \url{https://github.com/felmoreno1726/Expresso-AI}.} to better understand and explain decisions made by video-based deep learning models, specifically in the context of ADD. By training deep neural networks on facial video clips and analyzing sample attributions in relation to facial expressions and other self-interpretable lower-level face semantics, we aim to 
% enhance the interpretability of depression diagnosis models through quantitative measure.
identify quantifiable interpretation methods for video-based architectures by correlating these attributions with depression severity level, facial expression and facial regions.

\section{Related Work}
Interpretability of machine learning models focuses on approaches like prediction visualization, justification, and interpretable models, with Explainable AI (XAI) techniques aiming for human-like explainability \cite{biran2017explanation, miller2019explanation}. Visualizing parts of an input contributing to predictions in deep learning architectures has gained attention for interpretability \cite{hohman2018visual, nguyen2019understanding, fong2019explanations}, but instance group visualization, especially for video frame sequences, has received less attention \cite{li2020comprehensive,li2020depression}. Attribution-based visualization aids in model evaluation, despite current methods like instant inspection and ground truth evaluation being subjective due to their reliance on human observation \cite{bylinskii2018different}.

% ADD has attracted interest from researchers across various fields, as it holds the potential to provide objective measures for identifying and assessing depressive symptoms. With the emergence of deep learning techniques, numerous recent studies have employed different deep learning network architectures to model depression diagnosis using behavioral information, images, videos, and audio. In comparison, traditional machine learning techniques, such as Support Vector Machines (SVM), have been widely used in depression diagnosis over the past decade.
ADD has gained interest among researchers for its potential to objectively identify and assess depressive symptoms, especially with the emergence of deep learning techniques and various data types (images, videos, and audio). 
% Traditional machine learning techniques, such as Support Vector Machines (SVM), have also been widely used in depression diagnosis.
% However, most studies investigating ADD neither report interpretations of their models nor discuss the features or patterns that contributed to their performance. This may be due to the complexity of the extracted features or the techniques used for modeling. 
% However, the complexity of these techniques makes it difficult to interpret the features contributing to their performance improvements.
% For example, \cite{8501575} captured the microstructure of facial appearance by extracting local binary patterns (LBP) from three orthogonal planes of video frames. The Fisher vector was then used to cluster the features before training a support vector regression for depression severity modeling. Despite the high performance of their model, such approaches are difficult to interpret given the complexity of processing the features. In \cite{bhatia2017multimodal}, bag-of-words features from both speech and facial expressions were used to classify melancholic and non-melancholic depressed patients. Given the process of extracting bag-of-words, it is difficult to trace the behaviors that contributed to the classification models. More specifically, \cite{al2018depression} focused on eye blinks and blink duration as indicators for depression using several traditional classifiers for comparison, but no further analysis was provided to quantify the features. 
Various studies have used complex methods like extracting local binary patterns, bag-of-words features, and eye blink indicators to model depression severity and classify patients \cite{al2018depression}. However, interpreting these approaches is difficult due to the complexity of processing the features and the lack of further analysis to quantify them.

% Despite deep learning network architectures often outperforming traditional techniques, the complexity of deep learning techniques makes it difficult to interpret the features that contribute to such performance improvements. 
% A wide range of neural network architectures have been employed for depression modeling. 
% For example, using behavioral information such as reported mood, action, medications, and sleeping patterns, a recurrent neural network (RNN) was trained, which could forecast severe depression mood up to two weeks in advance \cite{Suhara2017DeepMood}. Another example of deep learning utilization for depression detection is presented in \cite{Alhanai20181716}, where audio and text features are extracted from virtual agent-human interaction interviews. A long-short-term memory (LSTM) neural network model is built using audio and text features. Furthermore, various neural network architectures have been employed for depression modeling, such as CNN \cite{pampouchidou2017quantitative}, C-CNN \cite{haque2018measuring}, Deep Autoencoding \cite{dibekliouglu2017dynamic}, multi-scale temporal dilated CNN \cite{yin2019multi}, DCNN \cite{7812588}, hybrid deep CNN \cite{yang2018integrating}, DepArt-Net \cite{du2019encoding}, and 3D-CNN \cite{de2020encoding}. %However, none of these studies provided interpretation or insight into what contributed to their model performances. 
% Deep learning techniques, despite often outperforming traditional methods, are difficult to interpret due to their complexity. 
Deep learning techniques outperform traditional methods but are complex to interpret.
Various neural network architectures, such as RNN \cite{Suhara2017DeepMood}, LSTM \cite{Alhanai20181716}, C-CNN \cite{haque2018measuring}, multi-scale temporal dilated CNN \cite{yin2019multi}, DCNN \cite{7812588}, hybrid deep CNN \cite{yang2018integrating}, DepArt-Net \cite{du2019encoding}, and 3D-CNN \cite{de2020encoding}, have been employed for depression modeling, making interpretation challenging.

Most ADD studies do not provide direct interpretation of their model performance, with only a few attempting to explain performance through statistical analysis or by linking with psychology literature \cite{cohn2018multimodal}. For instance, \cite{stepanov2018depression} modeled modalities and groups of features individually, finding that speech behaviors had higher performance compared to language and visual cues. In \cite{anis2018detecting}, histograms were used to analyze facial landmarks and 3D head motion features, revealing that velocity and acceleration of facial movement strongly mapped onto depression severity. The DepressNet model by \cite{zhou2018visually} visualized activation maps, showing the eye region to be the most indicative of depression severity, but lacked clarity on facial movement dynamics. The study by \cite{de2019depression} used attention maps in a ResNet-50 architecture to show differences in depression levels from the face, but did not directly explain the contributing dynamics and features. Lastly, \cite{song2020spectral} used CNN to model depression severity with handcrafted features, indicating action units around the eyes and mouth as the most indicative of depression levels, but did not provide quantified conclusions or deeper insights into model performance.

In our previous work, we made an effort to align our traditional depression detection models with statistical analysis of the extracted features and with cross-cultural generalizations in order to provide insight into the model performances and to link the machine learning literature with psychology literature \cite{Alghowinem2015}.
% In \cite{Alghowinem2015}, we investigated eye activities (e.g., blinking, iris movement) as indicators of depression, where we found that the average duration of eye blink and the average openness of the eyelids contributed to the high performance of the model. Similarly, head orientation, including its frequency and speed, were good indicators for depression and helped in its modeling \cite{Alghowinem2015}. 
However, these efforts do not provide direct insight into the model performance.

In this work, we aim to explain and interpret the predictions of ADD models by correlating the results with the predictions of less complex and more intuitive models as ground truths. These models include those that detect facial keypoints and facial expressions. Facial landmarks are predicted as heatmaps of the most likely region of localization, making them visually interpretable by default. The Action Unit model employs simple SVM and SVR modeling to detect expressions and calculate their intensity from facial appearance and geometric features. 

\section{Experimental Settings}
\subsection{Dataset}
We used the benchmark Depression Recognition Sub-challenge of the Audio/Visual Emotion Challenge (AVEC) 2014 \cite{valstar2014avec} for modeling depression for interpretation. The dataset consists of video recordings of 150 participants while performing two tasks: reading aloud and responding to questions in a freeform manner, such as discussing a favorite dish or a childhood memory. Though the AVEC (DAIC-Woz) 2019 dataset \cite{ringeval2019avec} provides audio-visual recordings, it has not been extensively utilized for visualization and interpretation. Consequently, we employ the AVEC 2014 dataset for comparisons with previous literature.

The dataset contains 300 video recordings, provided as separate videos for each participant, and divided into three dataset splits: Training, Development, and Testing. In our experiments, we merge the Training and Development sets for use as training data, and report performance based on the evaluation of the Testing set.

The videos, recorded at variable frame rates and resolutions, were later re-sampled to 30 frames per second (fps) and 640x480 pixels. The data labels correspond to the Beck Depression Inventory-II (BDI-II), a frequently used self-reported metric consisting of 21 multiple-choice questions. 

\subsection{Data Preprocessing}
For each video, the face is detected using the Single Shot Scale-Invariant Face Detector (S3FD or SFD) \cite{zhang2017s3fd}, a CNN-based detector capable of detecting faces at multiple resolutions and scales. Frames with unreliable face detection (confidence score $< 0.75$) are skipped. The detected face is fed to the Face Alignment Network (FAN) \cite{bulat2017far}, which extracts facial landmarks for each frame. The tightest square around the face is cropped out without alignment.

The crops of each individual region are up-sampled using LANCZOS4 interpolation, a robust method for small-scale crops \cite{Turkowski_turkowskifilters}. The dimensions are up-sampled to either 112x112 or 150x150, depending on the input model. Then, crops of each individual region are concatenated together into a video.
% The coordinates of the face bounding box and facial landmarks are also stored, along with the positions where a face was not detected reliably, in an array of 0/1 valid/invalid frames.

A sample is taken from the pre-processed data. In our experiments, the video clips fed into the model have lengths of 16, 32, or 64 frames. The frames of the clip cover a \textit{window size} of 256 or 512 frames. We perform input dilation, i.e., one frame is taken from every 8 or 16 frames. The clips overlap with each other, covering 40-60\% of the window.
We perform regularization by applying horizontal flips to training samples with a probability of 0.5.

\subsection{Network Architectures}
%explanation of depth vs. width Architectures
%explanation on base models experiments for transfer learning
Since we wanted to conduct the interpretation framework on the highest performing network architectures, we explored several ones.  
ResNet, or Residual Network, is a popular choice for various computer vision tasks due to its strong performance and ability to efficiently handle deep architectures. Therefore, in this work, we opt for ResNet-type architectures fine-tuned on Action Recognition datasets, with different variations as follows:

\begin{itemize}[leftmargin=*]
    % \item \textbf{Facenet}. An Inception ResNet pre-trained on either VGGFace2 or Casia-Webface. This model serves as our baseline and aims to reproduce the results from \cite{zhou2018visually} but with a more modern Inception architecture.
    % \item \textbf{Facenet Reverse Mixed Convolutions (RMC)}. A custom architecture made up of a 2D Facenet model with 3D convolutions stacked on top. The 2D model was pre-trained on the Casia-Webface dataset.
    % \item \textbf{Mixed Convolutions (MC)}. Deep Residual Networks with 3D convolution filters for low-level features and 2D convolution filters for high-level features. Models pre-trained on Kinetics-400.
    \item \textbf{ResNet 3D (R3D)}. Deep Residual Networks (ResNet) with 3D convolution filters. Pre-trained on Kinetics-400, Kinetics-700, and/or Moments in Time (MiT) datasets.
    \item \textbf{2-Plus-1 Dimensions ((2+1)D)}. Deep Residual Network which factorizes spatio-temporal (3D) convolution into a 2D spatial convolution and a 1D temporal convolution. Models were pre-trained on Kinetics-400, Kinetics-700, MiT, and/or Sports-1M.
\end{itemize}

% \subsection{Optimization}
% These networks were pre-trained with the following parameters: 
These pre-trained networks had these parameters:
\begin{itemize}[leftmargin=*]
    \item \textbf{Loss Function. } We use a L2 loss, also known as Mean Squarred Error (MSE) loss. This loss is fitting to the regression problem. 
    
    \item \textbf{Optimizer. } We use Stochastic Gradient Descent (SGD) with momentum and weight decay.
    
    \item \textbf{Learning Rate Scheduling. } We use Cosine Annealing. This in combination with SGD, is called SGD With Warm Restarts.
    
    \item \textbf{Freezer. } We developed an object called Freezer, which takes in a model, freezes all but the last layer, and progressively unfreezes the layers of the model top to bottom. Research shows that when fine-tuning transfer learning models, it is often desirable to train the top layers longer than the bottom layers. 
    
    \item \textbf{Gradient Clipping. } To avoid exploding and vanishing gradient problems as well as increasing regularization, we clip gradients during backpropagation to a gradient norm of 1.0. 
    
    \item \textbf{Regularization Noise. } To reduce overfitting, we apply random Gaussian noise to the target depression level during the training routine. The noise applied has mean 0, and standard deviation 0.5, 1.0, 1.5 or 2.0.
\end{itemize}

\subsection{Evaluation}
%- Clustered: (MSE), RMSE, MAE.
To measure the model's recognition performance, two evaluation metrics are used: Mean Absolute Error (MAE) and Root Mean Squared Error (RMSE). These metrics were selected for comparisons with prior literature on this dataset. The MAE and RMSE are calculated for each video sample in the dataset using the following equations:
These evaluation metrics help quantify the model's performance by measuring the difference between the predicted depression scores and the actual (ground truth) depression scores for each video sample in the dataset. 
% Since each video is segmented for modeling, the predicted depression score for a video is defined as the average of the predicted scores of all the segmented video clips from that video.
The predicted depression score for a video is determined by averaging the predicted scores of all its segmented clips.
% \begin{equation}
%     \textbf{MAE} = \frac{1}{N} \sum_{i=1}^{N} |y_i - \hat{y_i}|
% \end{equation}

% \begin{equation}
%     \textbf{RMSE} = \sqrt{\frac{1}{N} \sum_{i=1}^{N} (y_i - \hat{y_i})^2 }
% \end{equation}

% where N is the number of video samples in the dataset split (either train or validation), $\hat{y_i}$ and $y_i$ are the ground truth and inferred prediction for video segment $i$, respectively. 

\section{Interpretation of the ADD Model}
Early efforts in machine learning interpretability have concentrated on visualizations. Past research has demonstrated the effectiveness of occlusion techniques, activation-based methods, and backpropagation methods when applied to images. However, when it comes to videos, the majority of visualizations have employed activation-based methods (e.g. 2D-CNNs \cite{gan2015devnet}), with only a few occlusion methods (e.g. Spatio-Temporal Extremal Perturbation (STEP) \cite{litowards}) and hardly any backpropagation methods (e.g. DeepLift \cite{bach2015pixel}). More sophisticated techniques might be capable of quantifying model decisions beyond just offering qualitative visual explanations. 
We suggest that the fundamental properties of backpropagation methods such as DeepLift could provide satisfactory explanations that go beyond the visual domain and encompass the temporal aspect of explanations as well.
We use the DeepLift algorithm with Rescale Rule to compute attribution maps, which is a computationally efficient method with a conservation property. 
% That is, the total `relevance' contribution that flows into a neuron, leaves that neuron  \cite{bach2015pixel}. 
This method may provide explanations in the temporal realm, as well as the visual realm, for depression modeling, as explained in the following subsections.

\subsection{Attribution Generation}
The DeepLift algorithm is a model-agnostic interpretability method used to explain the predictions of deep learning models. To identify which features contribute most to the model's predictions, DeepLift assigns relevance scores to each input feature, which represent their contribution to the model's output \cite{bach2015pixel}.
It compares the activations of each neuron to a reference activation, which is typically chosen as the activation that occurs when the input is a neutral or baseline input. 

The Rescale Rule is an essential part of the DeepLift algorithm, where it distributes the relevance scores across the input features while maintaining the conservation property. The Rescale Rule ensures that this property holds true while rescaling the relevance scores to maintain their relative importance. That is, the sum of the relevance scores of all input features is equal to the model's output. 

In our research, we utilize the DeepLift algorithm combined with the Rescale Rule to calculate attribution maps. We selected this technique due to its low computational cost and the conservation property, mentioned above. The highest relevance in the algorithm's back-propagation process is the model's prediction score, which is then propagated all the way down to the model's inputs, generating relevance (or attribution) maps that satisfy the following equation:

\begin{equation} \label{eq:conservation}
        f(x) = \sum_{p, t}^d R_{p, t}
\end{equation}

In this equation,  $f$ represents the function computed by the model, $f(x)$ is the output generated based on input $x$ and $R_{p, t}$ denotes the relevance of pixel $p$ in frame $t$. We presume that other algorithms that fulfill the conservation property, such as Layer-wise Relevance Propagation (LRP) or Integrated Gradients, will provide interpretations similar to those obtained using DeepLift.

In this context, the DeepLift algorithm is applied to a 50-layer R(2+1)D model, which is a type of deep learning model used for video recognition tasks. For each sample in the testing set, attributions are computed using DeepLift, resulting in an attribution map with dimensions (T, C, W, H) = (16, 3, 112, 112) for each sample, where T is the time axis; C is channels (RGB); W is width; and H is height.

By applying the DeepLift algorithm to the R(2+1)D model, we can generate attribution maps for each video sample that reveal the importance of each pixel in the video frames (across time and color channels) to the model's predictions. 
% This helps to better understand and interpret the decision-making process of the model.
This aids in comprehending and interpreting the model's decision-making process.

\subsection{Relevance Pooling}
For model interpretability, Relevance Pooling aims to group input features into broader clusters based on their relevance scores or importance to the model's predictions. By focusing on these coarser groups of input features rather than individual spatio-temporal pixels, it concentrates on more significant patterns in the input data. 

Our analysis focuses on clustering relevance based on the conservation property, as shown in the equation \ref{eq:conservation}. Instead of examining every single input feature (spatio-temporal pixel), we are more interested in broader groups of input features. The conservation property over the input space can be represented by the following equation:

\begin{equation}
    f(x) \approx \sum_{t \in T} \sum_{c \in C} \sum_{w \in W} \sum_{h \in H} R_{t,c,w,h}
\end{equation}

In this equation, $R_{t,c,w,h}$ represents the relevance of each pixel in the video frames (across time and color channels) to the model's predictions.

\subsection{Video Attribution Maps}
\label{subsec:video_attributions}
% Relevance scores, or attributions, are grouped into coarser feature sets to provide better explanations and understanding of the model's behavior.
Relevance pooling is first performed along the channels axis (RGB color channels). For each attribution map, the relevance scores of the RGB channels of each spatio-temporal pixel are added together. As a result, a group of feature sets with dimensions (16, 112, 112) is obtained for each input. 

For the video, these aggregated feature sets are then used for heatmap visualizations. Heatmaps help to visually represent and identify the most important areas in the input data (in this case, video frames) that significantly contribute to the model's predictions. This is performed using the following equation:

\begin{equation}
    \begin{aligned}
     f(x)  & \approx \sum_{t \in T} \sum_{c \in C} \sum_{w \in W} \sum_{h \in H} R_{t,c,w,h} \\
     & = \sum_{t \in T}  \sum_{w \in W} \sum_{h \in H} (\sum_{c \in C} R_{t,c,w,h}) \\
     & = \sum_{t \in T}  \sum_{w \in W} \sum_{h \in H} R^C_{t, w, h}
    \end{aligned}
\end{equation}
        
\subsection{Temporal Attributions}
The goal is to create a vector of temporal attributions ($R^T$) that simplifies the analysis of the model's behavior by focusing on the most critical patterns in the input data over time.
By pooling the relevance for each frame into a single attribution score, we focus the interpretation on the most important temporal patterns in the input data.
This vector of temporal attributions $R^T$ satisfies:
    \begin{equation}
    \begin{aligned}
         f(x)  & \approx \sum_{t \in T} (\sum_{c \in C} \sum_{w \in W} \sum_{h \in H} R_{t,c,w,h}) \\
         & = \sum_{t \in T}  R^T_t
    \end{aligned}
    \end{equation}

\subsection{Region-wise Attributions}
To objectively interpret the model in regard to facial parts, we calculated attributions (relevance scores) around specific regions of interest in the face, such as mouth, right eyebrow, left eyebrow, right eye, left eye, nose, jaw, besides the whole face. These regions are identified using pre-computed facial landmarks.
In this approach for pooling relevance, the conservation property is not satisfied. However, in this region-wise analysis, attributions outside the regions of interest are discarded, resulting in the conservation property not being maintained.

\subsection{Video Heat Maps Visualizations}
Video heat maps visualizations highlight the most relevant regions in a video by normalizing across the spatio-temporal volume of the video. These video heat map visualizations could be used to identify important spatio-temporal features in videos of individuals being assessed for depression, which allow for a better understanding and interpretation of the modeling process. As discussed, this could potentially lead to improved models for detecting and diagnosing depression, as well as insights into the specific visual cues that the models are using to make their predictions.

To our knowledge, no previous work has attempted to visualize back-propagation-based attribution maps for video modeling. Interestingly, we found that normalizing across the spatio-temporal volume produced very smooth maps of relevant regions. This is quite different from image visualizations, which are known to be highly pixelated in areas of interest.

For each video clip attributions, we normalize across the spatio-temporal volume using a cumulative sum that maps 98\% of the attributions to the desired range: positive (0, 1), negative (-1, 0), all (-1, +1), or absolute value (0, 1). We let 2\% of the outliers fall outside the range. We present visualizations of the absolute value attributions normalized to the range (0, 1). These visualizations are designed to display the most important regions in brighter colors.

\subsection{Image Grids Visualizations}
Image grids model visualizations identify and analyze the most important video frames that contribute to the modeling results. In this method, video-clip frames with the highest and lowest attributions in $R^T$ are extracted. The highest attributions represent the frames that contributed the most to a depression diagnosis, whereas the lowest attributions represent frames that contributed the most to a non-depressed diagnosis.

By visualizing the individual attributions for each frame, we can gain insights into the specific features and patterns that the depression model is using to make its predictions. These visualizations are called saliency maps, and they tend to be highly pixelated, highlighting the most relevant pixels in each frame.
This method would derive the factors that the ADD model is using to diagnose depression, where visual cues are analyzed from the most influential frames. Moreover, this method can also help identify potential biases or issues in the model, ensuring that the model's predictions are reliable and accurate.

\subsection{Action Unit Cross Correlation Analysis}
To provide an objective and interpretable measure of our model's performance, we analyzed the relationship between Action Units (AUs) and the model's Temporal Attributions (TAs) signal in the context of depression recognition. AUs are facial muscle movements that can be used to describe facial expressions. Therefore, correlation analysis of AUs and the model's TAs would provide insight into the model's performance. 
Action Units (AUs) are extracted from each of the AVEC 2014 original recordings using the OpenFace analysis toolkit \cite{baltrusaitis2018openface}. AUs are extracted using dynamic normalization, which post-calibrates the predictions for each video by finding the person's median facial expression to correct over- and under-prediction of AUs \cite{baltruvsaitis2015cross}.

To quantify the relationship between the temporal presence of AUs and the model's TAs, the Kendall Tau correlation coefficient is calculated. This coefficient measures the correlation between two variables, with scores ranging from -1 (negative correlation) to +1 (positive correlation). Tau is scale-invariant, making it suitable for comparing attributions across different depression classes.

However, due to the short duration of the video clips (16 frames), the correlation coefficients calculated for individual clips might not be reliable. An AU may be present throughout the entire clip or not present at all, leading to no apparent correlation between the AU and the model's decision. To address this issue, the AU cross-correlation scores are averaged across all clusters, providing a more comprehensive understanding of the relationship between AUs and the model's decision-making process, considering the relatively short duration of the clips.

\section{Results}
%the table of results from your thesis
%explanation and comparisons of the results 
\subsection{Architecture Selection Results}
\begin{table}[t]
\centering
\caption{Performance Comparison of single face ADD in the AVEC 2014 Dataset with our fine-tuning results.}
\label{table:merged_performance}
\begin{tabular}{l|c|c|c|c}
\hline\hline
Method           & MAE & RMSE &  &  \\ \hline\hline
Baseline \cite{valstar2014avec}        & 8.86                    & 10.86 & & \\
UUIMSidorov \cite{sidorov2014emotion}    & 11.20                   & 13.87 & & \\
InaoeBuap \cite{perez2014fusing}      & 9.35                    & 11.91 & & \\
Brunnel \cite{jan2014automatic}   & 8.44                    & 10.50 & & \\
Zhu \cite{zhu2017automated}              & 7.47                    & 9.55  & & \\
BU-CMPE \cite{kaya2014ensemble}          & 7.96                    & 9.97  & & \\
Jan \cite{jan2017artificial}           & 6.68                    & \textbf{8.04} & & \\
DepressNet-Full \cite{zhou2018visually}   & \textbf{6.60}                   & 8.88  & & \\
% DepressNet-Top    & 6.86                   & 9.81  & & \\
% DepressNet-Central & 6.39                  & 8.55 & & \\
% DepressNet-Bottom   & 7.51                 & 9.37  & & \\
% MR-DepressnetNet & \textbf{6.21}           & 8.39  & & \\ 
\hline

\textbf{Ours}   &                  &   & Pre-train Dataset & Depth \\\hline
R3D          & 7.430          & 9.012 & Kinetics-400       & 18    \\ 
R(2+1)D      & 6.820 & 8.640 & Kinetics-400       & 18    \\ 
Wide ResNet  & 6.984          & \textbf{8.529} & Kinetics-400       & 50    \\ 
R3D          & 7.228          & 9.278 & Kinetics-400       & 101   \\ 
R3D          & 6.999          & 8.903 & Kinetics-700       & 101   \\ 
R3D          & 6.700 & 8.694 & Kinetics-700 + MiT & 101   \\ 
R(2+1)D      & 7.027          & 8.687 & Kinetics-700       & 50    \\ 
R(2+1)D      & \textbf{6.626} & 8.543 & Kinetics-700 + MiT & 50    \\ \hline\hline
\end{tabular}
\end{table}

To select the best performing model architecture for predicting depression scores in videos, we fine-tune a few ResNet architectures trained in different datasets as backbone, with different depth levels (see Table \ref{table:merged_performance}).
The models developed in this study outperform previous single-face ADD results, with the current benchmark held by the DepressNet-Full model. 
Based on the results from Table \ref{table:merged_performance}, R(2+1)D trained on both Kinetics-700 and MiT produced the highest results.
Therefore, interpretation analyses of the R(2+1)D model would provide more accurate and reliable predictions, paving the way for improved understanding and treatment of depression.

\subsection{Interpretation and Visualization}
%%%% CORRECT SEVERE, moderate, mild, minimal           
\begin{figure}[t]
    \centering
    \begin{subfigure}{\linewidth}
    \scalebox{0.9}{
        \includegraphics[width=\linewidth]{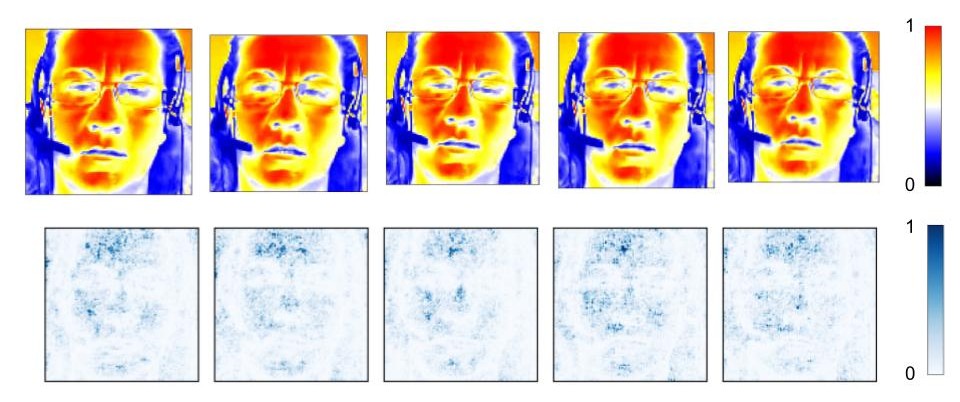}
        }
        \caption{Top attribution frames of severely depressed}
        \label{fig:video_visualizations_severe_correct}
    \end{subfigure}

    \begin{subfigure}{\linewidth}
    \scalebox{0.9}{
        \includegraphics[width=\linewidth]{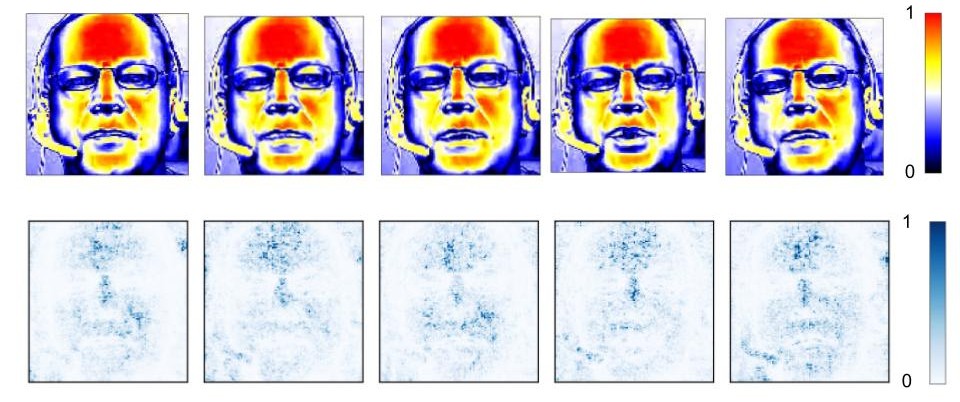}
        }
        \caption{Top attribution frames of moderately depressed}
        \label{fig:video_visualizations_severe_correct}
    \end{subfigure}

    % \begin{subfigure}{\linewidth}
    % \scalebox{0.9}{
    %     \includegraphics[width=\linewidth]{./figures/mild_top_grid.jpg}
    %     }
    %     \caption{Top attribution frames of mildly depressed}
    %     \label{fig:video_visualizations_severe_correct}
    % \end{subfigure}
    
    \begin{subfigure}{\linewidth}
    \scalebox{0.9}{
        \includegraphics[width=\linewidth]{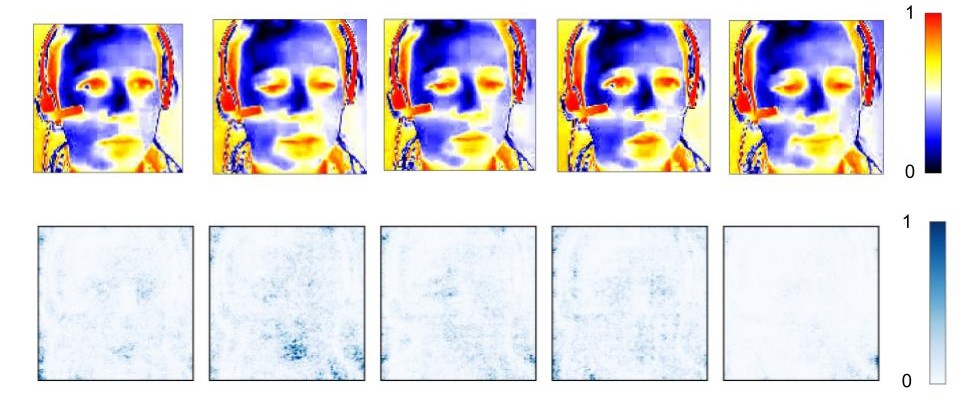}
        }
        \caption{Top attribution frames of minimally depressed}
        \label{fig:video_visualizations_minimal_correct}
    \end{subfigure}
    
    \caption{Visualization of the model's top attribution frames, showing examples of correctly estimated results from different level of depression severity.}
    
    \label{fig:vis_correct}
\end{figure}
Fig. \ref{fig:vis_correct} presents examples of attributions and heat maps from the top contributing frames to the model's results when it correctly predicts depression severity levels. It can be observed that the model follows specific patterns in its diagnosis.

For instance, the regions of the forehead exhibit higher attributions in severe and moderate depression levels compared to the minimal level. This suggests that tense frowns and eyebrow movements have more significant attributions in these cases. Similarly, the mouth regions display higher activations in minimal depression levels than in moderate and severe levels.

However, a notable issue is the high attribution scores for headsets and microphones in the video visualizations. This is likely due to a bias in the training set, where participants with headsets and microphones are more inclined towards depressive labels. This bias adversely impacts the dataset's generalization and affects the model's performance.

\begin{figure}[t]
    \includegraphics[width=\linewidth]{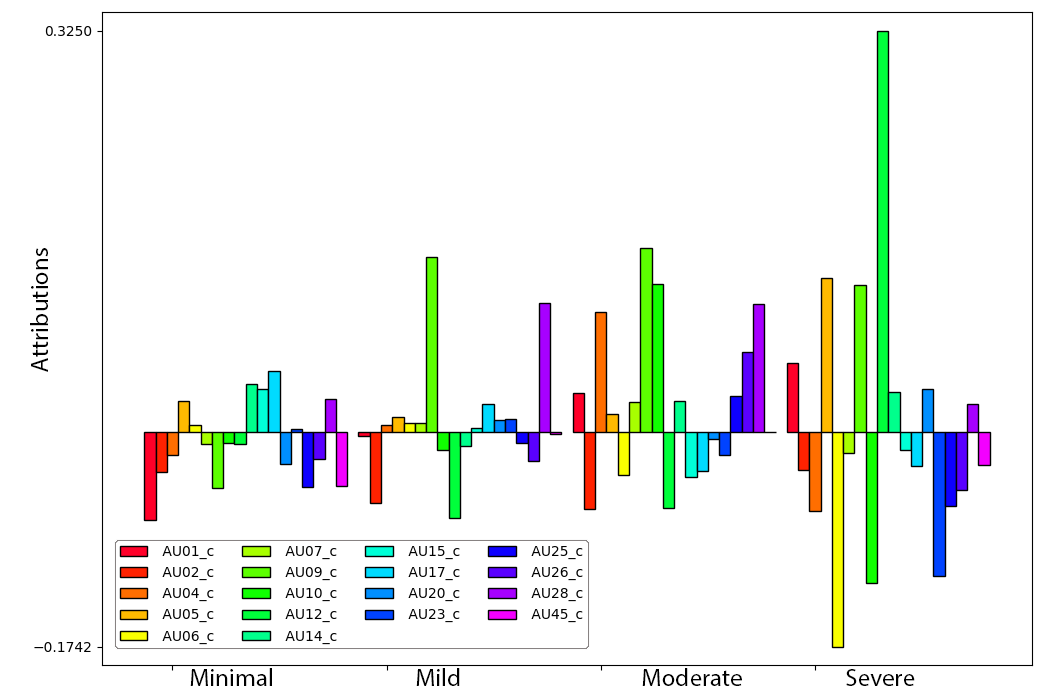}
    \caption{Cluster analysis of action unit cross correlations on correctly predicted samples.}
    \label{fig:Cluster_FAU}
\end{figure}

These qualitative observations call for a more objective and reliable assessment and interpretation of the model's performance. 
To objectively assess these attributions, a cross-correlation analysis is performed with specific regions and Action Units (AUs) (see Fig. \ref{fig:Cluster_FAU}). This analysis helps to better understand the relationship between facial expressions, as represented by AUs, and the model's attributions, providing a more accurate and reliable interpretation of the decision-making process.

\subsection{Action Unit Cross Correlation Analysis}
    
%%% AU CROSS-CORRELATION ANALYSIS
% \begin{figure}[t]
%     \centering
%     \includegraphics[width=\linewidth]{./figures/overall_au_c_attribution_correlations_attributions.png}
%     \caption{Cluster analysis of overall action unit cross correlations.}
% \end{figure}
    
% \begin{figure}[t]
%     \includegraphics[width=\linewidth]{./figures/overestimated_au_c_attribution_correlations_attributions.png}
%     \caption{Cluster analysis of action unit cross correlations on overestimated incorrectly predicted samples.}
% \end{figure}
    
% \begin{figure}[t]
%     \includegraphics[width=\linewidth]{./figures/underestimated_au_c_attribution_correlations_attributions.png}

%     \caption{Cluster analysis of action unit cross correlations on underestimated incorrectly predicted samples.}
% \end{figure}
    
    % Looking at the correct action unit correlation clusters, we can see the action units which are positively correlated with each cluster. Let's first look at the correctly predicted clusters. We see that AU9 (nose frowning) is and indicator of depression and is the most correlated with severe depression and is also present in mild and moderate clusters. It's also good to note AU 4 and AU10 (frowning) correlated with the moderate depression cluster. Also very interestingly, we see that the most negatively correlated AUs with depression are AU6 and AU12 which together make a Ducheness Smile (or authentic smile) an joy or happiness. 
Examining the Action Unit correlation clusters provides insights into the AUs that are positively correlated with each depression severity group. Focusing on the correctly predicted results, we observe that AU9 (nose wrinkling) is an indicator of depression severity and is most correlated with severe depression while also being present in mild and moderate clusters. It is also worth noting that AU4 (brow furrowing) is correlated with the moderate depression cluster, suggesting a connection between frowning and depression severity. These results are aligned with depression facial expression literature, where depressed individuals express anger (AU4) and disgust (AU9) \cite{pampouchidou2017automatic}. 

Interestingly, the AUs most negatively correlated with depression are AU6 (cheek raising) and AU12 (lip corner pulling), which, when combined, form a Duchenne Smile (or authentic smile) and represent joy or happiness. This finding is supported by the literature, and therefore, confirms that the model was able to find the association between specific facial expressions and different depression severity levels, highlighting the importance of understanding and interpreting the decision-making process of the model.

By identifying these correlations between AUs and depression severity, the research validates the model's ability to recognize patterns in facial expressions that are indicative of various depression levels. This understanding can contribute to improving the model's performance and generalization, ultimately leading to more accurate, interpretable and reliable detection and diagnosis of depression.

\subsection{Clustering Analysis}  
%%%% REGION-WISE ANALYSIS
% \begin{figure}[t]
%     \centering
%     \includegraphics[width=\linewidth]{./figures/overall_region-wise_attributions.png}
%     \caption{Cluster analysis of overall region-wise summaries}
%     \label{fig:overall_region}
% \end{figure}
    
\begin{figure}[t]
    \centering
    \includegraphics[width=0.9\linewidth]{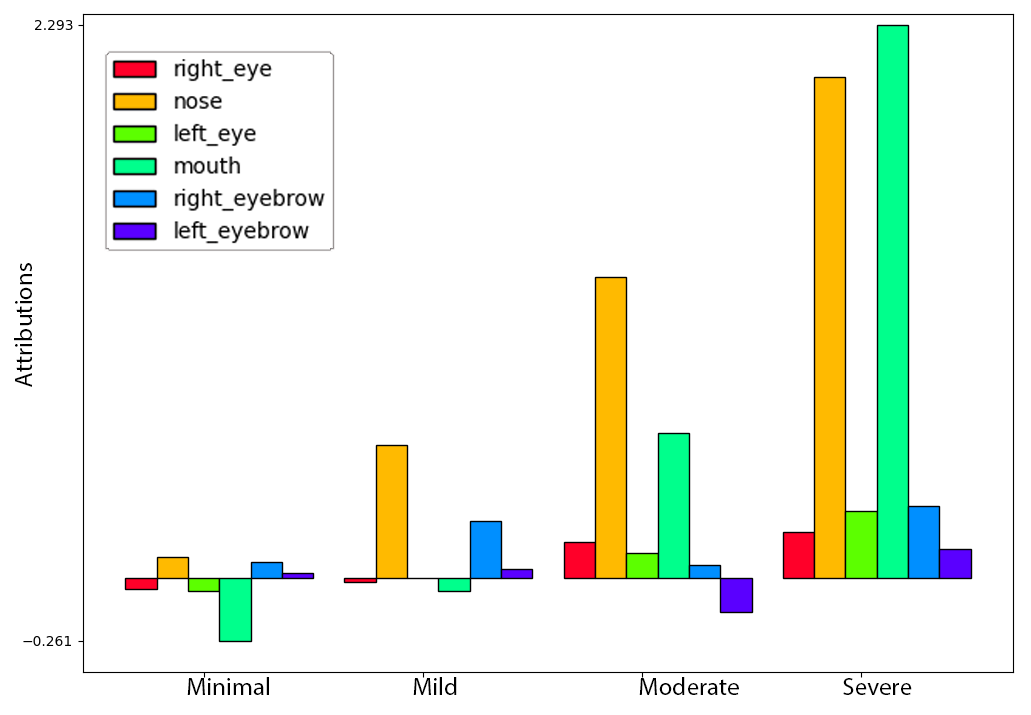}
    \caption{Region-wise analysis cluster of correctly predicted samples.}
    \label{fig:correct_region}
\end{figure}

% \begin{figure}[t]
%     \includegraphics[width=\linewidth]{./figures//overestimated_region-wise_attributions.png}
%     \caption{Region-wise analysis cluster of overestimated incorrectly predicted samples.}
%     \label{fig:overestimated_region}
% \end{figure}
    
% \begin{figure}[t]
%     \includegraphics[width=\linewidth]{./figures/underestimated_region-wise_attributions.png}

%     \caption{Region-wise analysis cluster of underestimated incorrectly predicted samples.}
%     \label{fig:underestimated_region}
% \end{figure}
    % Our findings for region-wise analysis are summarized in Fig. \ref{fig:correct_region}.
    % Looking at the region-analysis we also note the high importance of the nose and mouth to the prediction decisions. We observe that the mouth attributions tend to be high in correctly labeled depressed (moderate and severe) participants, but low in those misclassified (underestimated). We speculatively attribute this low to participants having a high vocal expressivity. On the other hand, correctly classified non-depressed (minimal or mild) participants tend to have low nose attributions, but when incorrectly-labeled (overestimated) these tend to have high nose attributions. We speculatively attribute this to nose expressions such as AU09 (nose wrinkling). 

The region-wise analysis findings are summarized in Fig. \ref{fig:correct_region}. This analysis reveals the importance of specific regions to the model's prediction decisions, which would provide insights and interpretation of the model.

Examining the mouth attributions, we observe that they tend to be high in correctly labeled depressed (moderate and severe) participants, but low in those misclassified (underestimated). We speculate that this lower attribution may be due to participants having high vocal expressivity, which could impact the model's interpretation of their depression severity.

On the other hand, correctly classified non-depressed (minimal or mild) participants tend to have low nose attributions, but when incorrectly labeled (overestimated) these tend to have high nose attributions. We speculate that this may be attributed to nose expressions such as AU09 (nose wrinkling), which could be associated with higher depression severity by the model.

% Similarly, the attributions are positively correlated with left and right eye regions in moderate and severe depression levels, while it is negatively correlated in mild and minimal levels of depression. 
Likewise, a positive correlation exists between attributions in the left and right eye regions and moderate to severe depression levels, while a negative correlation is observed in mild and minimal depression levels.
The positive correlation of attributions in the eye regions with moderate and severe depression levels suggests that specific eye-related expressions or movements could be indicative of higher depression severity. These expressions may include reduced eye contact, increased blinking rate, or gaze aversion, which are known to be associated with depression \cite{Alghowinem2018Multimodal}.

On the other hand, the negative correlation of attributions in the eye regions with mild and minimal depression levels implies that different eye-related expressions or behaviors might be more prevalent in individuals with lower depression severity. These behaviors could include more consistent eye contact, more natural blinking rates, or more engaged gaze patterns that are associated with non-depressed or less depressed individuals.

These findings further validate the main research question by demonstrating the importance of understanding and interpreting the decision-making process of the model. Identifying the significance of specific facial regions and their attributions can contribute to improving the model's performance and generalization. By recognizing patterns in facial expressions and their corresponding attributions, the model can make more accurate and reliable predictions, ultimately leading to better detection and diagnosis of depression.

\section{Conclusion}

In this work, we aimed to modernize visual Automatic Depression Diagnosis (ADD) by employing the latest techniques and architectures in video modeling. Our findings demonstrated that video-based models outperformed single-image-based modeling for ADD tasks. We identified the best video architectures for the task and leveraged pre-trained models on large-scale video datasets to enhance our model's performance. As a result, we developed a more optimal model for single-face ADD.

While exploring video interpretation, we recognized the potential for innovation in using a back-propagation saliency map method, allowing us to quantitatively assess both temporal and regional importance. We performed face segmentation in regions of interest, such as eyes, eyebrows, mouth, and nose, and calculated the associated relevance to better understand the decisions made by ADD models. Additionally, we expanded our understanding to the temporal frame by analyzing attributions in the context of Action Units as frame semantics. A preliminary correlation analysis found associations between severe depression and expressions of disgust and contempt, as well as a negative correlation between severe depression and expressions of joy. Among other findings, we identified talkativeness and mouth expressiveness as indicators of non-severe depression. We were intrigued to discover significant relevance in the nose area, which may indicate the importance of nose-wrinkling, an expression of disgust, in detecting depression. These findings are consistent with each other and with the clinical depression literature.

Future work on the face interpretation framework should focus on a careful statistical analysis of the findings. In this work, we are pleased to have generated hypotheses on face expressions and regions, which may be validated a posteriori with scientific research. However, determining statistical significance may show that the framework's findings are self-validating. 

%%  ---- START: Added 2022-03-15 for ACII2022 onwards -----
\section*{Ethical Impact Statement}
The research presented in this work aimed to advance the field of ADD by employing state-of-the-art techniques and architectures in video modeling. While the development of more accurate and reliable ADD models has the potential to improve the detection and diagnosis of depression, it also raises several ethical considerations that must be addressed.

1. Bias and Fairness: In this research, we found high attribution scores for headsets and microphones in the video visualizations, which could be an artifact of the training set. To ensure fair and unbiased predictions, future work should focus on addressing these biases and examining the model's performance across diverse populations and settings.

2. Transparency and Interpretability: In this work, we presented several methods to analyze the model's performance and interpretability, such as Relevance Pooling, Video Heat Maps Visualizations, and Region-wise Attributions. Ensuring transparency and interpretability allows for a better understanding of the decision-making process and can help identify potential biases or issues that may need to be addressed.

3. Clinical Validation and Application: Before deploying ADD models in real-world settings, it is crucial to validate their performance and accuracy through clinical trials and evaluations. Ensuring that the models are reliable and accurate can prevent potential misdiagnoses and unnecessary distress for patients. Additionally, it is essential to consider how these models would be integrated into existing clinical practices and to provide adequate training and support for healthcare professionals using these tools.

%%  ---- END: Added 2022-03-15 for ACII2022 onwards -----

\bibliographystyle{IEEEtran}
\bibliography{references}

\end{document}